\documentclass[iicol]{sn-jnl}
\usepackage[numbers]{natbib} %
\usepackage{graphicx}%
\usepackage{multirow}%
\usepackage{amsmath,amssymb,amsfonts}%
\usepackage{amsthm}%
\usepackage{mathrsfs}%
\usepackage[title]{appendix}%
\usepackage{xcolor}%
\usepackage{textcomp}%
\usepackage{manyfoot}%
\usepackage{booktabs}%
\usepackage{algorithm}%
\usepackage{algorithmicx}%
\usepackage{algpseudocode}%
\usepackage{listings}%
\usepackage{float}%
\usepackage{subfigure}
\usepackage{caption}

\theoremstyle{thmstyleone}%
\theoremstyle{thmstyletwo}%

\theoremstyle{thmstylethree}%

\begin{document}

\title[Article Title]{Skeleton-to-Image Encoding: Enabling Skeleton Representation Learning via Vision-Pretrained Models}

\author[1]{\fnm{Siyuan} \sur{Yang}}\email{siyuany@kth.se}

\author*[2]{\fnm{Jun} \sur{Liu}}\email{j.liu81@lancaster.ac.uk}
\author[3]{\fnm{Hao} \sur{Cheng}}\email{chenghao@hebut.edu.cn}
\author[4]{\fnm{Chong} \sur{Wang}}\email{wang1711@ntu.edu.sg}
\author[4]{\fnm{Shijian} \sur{Lu}}\email{Shijian.Lu@ntu.edu.sg}
\author[1]{\fnm{Hedvig} \sur{Kjellstrom}}\email{hedvig@kth.se}
\author[4]{\fnm{Weisi} \sur{Lin}}\email{wslin@ntu.edu.sg}
\author[4,5,6]{\fnm{Alex} \sur{Kot}}\email{eackot@ntu.edu.sg}

\affil[1]{\orgname{KTH Royal Institute of Technology},  \orgaddress{\city{Stockholm}, \country{Sweden}}}
\affil[2]{\orgname{Lancaster University}, \orgaddress{\city{Lancaster}, \country{UK}}}
\affil[3]{\orgname{Hebei University of Technology},  \orgaddress{\city{Tianjin}, \country{China}}}
\affil[4]{\orgname{Nanyang Technological University},  \orgaddress{\city{Singapore}, \country{Singapore}}}
\affil[5]{\orgname{Shenzhen MSU-BIT University},  \orgaddress{\city{Shenzhen}, \country{China}}}
\affil[6]{\orgname{VinUniversity},  \orgaddress{\city{Hanoi}, \country{Vietnam}}}

\abstract{Recent advances in large-scale pretrained vision models have demonstrated impressive capabilities across a wide range of downstream tasks, including cross-modal and multi-modal scenarios. 
    However, their direct application to 3D human skeleton data remains challenging due to fundamental differences in data format. 
    Moreover, the scarcity of large-scale skeleton datasets and the need to incorporate skeleton data into multi-modal action recognition without introducing additional model branches present significant research opportunities.
    To address these challenges, we introduce Skeleton-to-Image Encoding (S2I), a novel representation that transforms skeleton sequences into image-like data by partitioning and arranging joints based on body-part semantics and resizing to standardized image dimensions.
    This encoding enables, for the first time, the use of powerful vision-pretrained models for self-supervised skeleton representation learning, effectively transferring rich visual-domain knowledge to skeleton analysis. 
    While existing skeleton methods often design models tailored to specific, homogeneous skeleton formats, they overlook the structural heterogeneity that naturally arises from diverse data sources.
    In contrast, our S2I representation offers a unified image-like format that naturally accommodates heterogeneous skeleton data.
    Extensive experiments on NTU-60, NTU-120, and PKU-MMD demonstrate the effectiveness and generalizability of our method for self-supervised skeleton representation learning, including under challenging cross-format evaluation settings. }

\maketitle

\section{Introduction}
\label{sec:intro}

\begin{figure*}[t]
    \centering
    \includegraphics[width=1\linewidth]{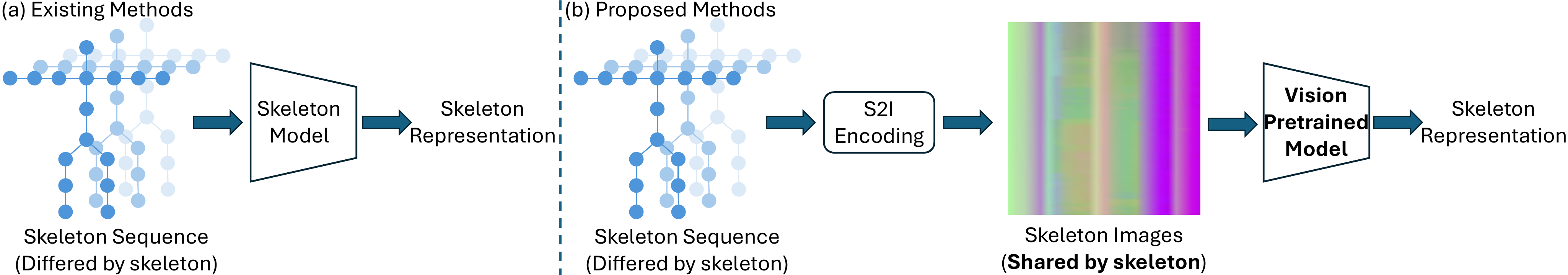}
    \caption{Overview of the existing methods and our proposed method. The existing methods train the skeleton models directly, while the proposed method converts skeleton data into image-like data and then train with the vision models, which can be initialized with pre-trained weights. }
    \label{fig:model_compare}
\end{figure*}

Recent years have witnessed the remarkable success of large-scale vision-pretrained models, such as Vision Transformers (ViTs)~\cite{dosovitskiy2021an}, Masked Autoencoders (MAE)~\cite{he2022masked}, and Vision-Language Models (VLMs)~\cite{ pmlr-v139-jia21b,Singh_2022_CVPR}, across diverse visual recognition and understanding tasks. 
These models leverage abundant image data to learn transferable representations and are increasingly adapted to other modalities, including depth maps~\cite{xia2024large,yang2024depth}, IR images~\cite{zhang2023pad,paranjape2025f}, video sequences~\cite{tong2022videomae, wang2023videomaev2, wang2022omnivl}, and even point clouds~\cite{wang2022p2p,zhang2022pointclip}.
A common strategy is to project non-image modalities into 2D formats, enabling direct use of image-pretrained models. While this works for dense 3D data like point clouds, applying similar projections to 3D skeleton data poses unique challenges. 
Skeletons are inherently sparse (only 15-30 joints per frame) and exhibit articulated structures with strong semantic relationships—unlike unstructured point clouds. Moreover, skeleton sequences span temporal dimensions critical for motion understanding, making naive 2D projections inadequate.

While directly applying vision models to skeleton data is non-trivial, skeleton-based representation learning remains fundamental for understanding human motion.
Skeleton data provides a compact, appearance-invariant, and high-level abstraction of human activities, making it particularly valuable for tasks such as action recognition, gait analysis, and human-computer interaction. 
Furthermore, as multi-modal action recognition gains traction, skeleton data serves as a complementary modality that can enhance robustness and interpretability when integrated with dense visual modalities such as RGB and depth.
However, the scarcity of large-scale annotated skeleton datasets and the incompatibility of skeleton structures with existing vision model architectures limit current methods' generalization ability across diverse tasks and scenarios.

To overcome these challenges, we introduce an approach that leverages pretrained Vision Models, such as MAE~\cite{he2022masked} and DiffMAE~\cite{wei2023diffusion}, for skeleton representation learning. 
This extends the use of vision models beyond 2D images to the 3D skeleton domain by transferring rich knowledge from large-scale image pretraining.
At the core of this approach is our proposed \textbf{Skeleton-to-Image Encoding (S2I)}, a novel representation method that reformats skeleton sequences into image-like representation compatible with vision models.
Specifically, the 3D joint coordinates $(x, y, z)$ are directly mapped to RGB channels, converting motion patterns into pseudo-images.
To ensure semantic consistency, we first partition skeleton joints into five body parts: torso, left arm, right arm, left leg, and right leg. 
These are then reordered by following the body part sequence, and within each part, joints are further sorted in a top-down manner based on their physical positions. 
We then stack these reordered joints across the temporal dimension, producing a spatial-temporal image-like representation of the entire skeleton sequence. 
Finally, the generated representation is resized to the standard image input size ($224 \times 224$), enabling seamless integration with vision model inputs.
As a result, our method enables, for the first time, the direct application of powerful pretrained vision models for self-supervised skeleton representation learning, effectively transferring rich visual domain knowledge to the skeleton domain without requiring task-specific architectural modifications. A visual comparison between existing skeleton pipelines and our approach is shown in Figure~\ref{fig:model_compare}.

Current skeleton-based methods are typically designed for homogeneous skeleton formats, relying on fixed joint definitions and dataset-specific architectures. Such designs limit their scalability and make it difficult to accommodate skeleton data with varying joint configurations, coordinate systems, or capture devices. As a result, these methods struggle in cross-format scenarios, where skeleton layouts differ significantly across datasets.
In contrast, our proposed S2I provides a unified and format-agnostic representation framework. By abstracting skeleton data into a consistent image-like structure, our method naturally supports joint training across multiple heterogeneous skeleton datasets.
This capability enables universal skeleton representation learning, where diverse skeleton formats can be leveraged together to enhance model generalization and capture richer motion dynamics.

Through extensive experiments on benchmark datasets, including NTU-60, NTU-120, and PKU-MMD, we demonstrate that our method achieves competitive performance in standard in-domain self-supervised skeleton representation learning. More importantly, S2I shows stronger advantages in transfer-oriented settings, especially cross-dataset and cross-format evaluation, where skeleton layouts, joint definitions, and data collection domains differ substantially. This robustness stems from the proposed unified image-like representation, which enables the model to benefit from large-scale vision-pretrained weights and, further, to leverage heterogeneous skeleton datasets for universal pretraining. These results indicate that S2I is particularly effective when transferring learned representations to new datasets and skeleton formats.


Our contributions can be summarized as follows:
\begin{itemize}
    \item We propose a novel pipeline that leverages vision-pretrained models and their weights for skeleton-based representation learning, bridging the modality gap between images and skeleton sequences.
    \item We introduce Skeleton-to-Image Encoding, a unified representation method that reformats sparse 3D skeleton data into image-like inputs, compatible with vision models and resilient to skeleton format variations. 
    \item We are the first to explore heterogeneous skeleton representation learning and propose a universal skeleton pretraining strategy, enabling cross-format learning across multiple skeleton datasets. 
\end{itemize}
  
\section{Related Works}

\subsection{Skeleton-based Action Recognition}
Skeleton-based action recognition has been widely studied due to the compact and appearance-invariant nature of skeleton sequences. Early deep learning methods mainly modeled skeleton dynamics with recurrent networks, convolutional networks, or graph-based networks. RNN-based methods employed temporal modeling to capture motion dependencies from joint sequences. For example, hierarchical RNNs were introduced to model body structures and temporal dynamics~\cite{du2015hierarchical}, while spatial-temporal LSTMs and attention-based LSTMs further exploited joint dependencies and discriminative temporal segments~\cite{liu2017skeleton,liu2016spatio,song2017end}.

CNN-based methods have also been widely explored for skeleton action recognition~\cite{du2015skeleton,ke2017new,li2017skeleton,li2018co,soo2017interpretable,wang2017scene}. Since CNNs operate on regular grids, these methods typically reorganize skeleton sequences into structured maps or image-like representations, or apply temporal convolutions to skeleton sequences. For example, early CNN-based approaches arrange joint coordinates over time into 2D skeleton maps, while other methods design translation- and scale-invariant mappings, enhanced skeleton visualizations, temporal convolutional representations, or co-occurrence feature learning frameworks. These studies show that skeleton data can be processed by convolutional architectures when represented in a suitable structured form. However, they are mainly designed for supervised action recognition with CNNs trained from scratch under fixed skeleton formats.

Inspired by the observation that the human skeleton naturally forms a topological graph, GCN-based methods have become a dominant paradigm for skeleton-based action recognition. ST-GCN~\cite{yan2018spatial} models skeleton sequences as spatial-temporal graphs and performs graph convolution over joints and frames. Subsequent methods improve graph construction and topology learning by introducing adaptive graph learning, bone streams, neural architecture search, lightweight shift operations, and channel-wise topology refinement~\cite{Shi_2019_CVPR_twostream,peng2020learning,cheng2020skeleton,chen2021channel}. More recent GCN variants further strengthen topology modeling through information bottleneck learning, hierarchical graph decomposition, and block-wise topology modeling~\cite{chi2022infogcn,lee2023hierarchically,zhou2024blockgcn}.

Transformer-based methods have also been introduced to capture long-range spatial-temporal dependencies in skeleton sequences. These methods usually tokenize joints, frames, or body parts and apply self-attention to model spatial-temporal relations beyond local graph neighborhoods. Recent approaches further improve attention modeling with skeleton-specific partitioning or efficient spatial-temporal designs~\cite{plizzari2021skeleton,do2024skateformer}. Although these GCN- and Transformer-based methods achieve strong performance on standard benchmarks, most of them rely on fixed joint layouts, predefined skeleton tokenization, or topology-aware architectural designs.

Different from prior supervised CNN-based skeleton models and skeleton-specific GCN/Transformer architectures, our work investigates whether large-scale vision-pretrained masked models can be directly adapted to skeleton representation learning through a simple image-like interface. Moreover, existing methods are mostly tailored to homogeneous skeleton formats with fixed joint numbers and joint orders, which limits their scalability to heterogeneous skeleton sources. In contrast, our S2I representation aims to provide a format-agnostic interface that can accommodate different skeleton layouts and support cross-format and universal skeleton representation learning.

\subsection{Self-Supervised Skeleton Representation Learning}
Self-supervised skeleton representation learning has become an important direction for reducing the reliance on costly manual annotations in skeleton-based action recognition. Existing methods mainly adopt contrastive learning or masked modeling to learn transferable motion representations from unlabeled skeleton sequences.

Contrastive methods generate different augmented views of the same skeleton sequence and encourage their representations to be close in the feature space~\cite{li20213d,guo2022contrastive,zhang2022contrastive,zhang2023prompted}. These methods often exploit spatial-temporal augmentations, cross-view consistency, or multi-stream skeleton modalities such as joint, bone, and motion to learn discriminative features. Later works further improve contrastive learning with stronger motion modeling, masked motion prediction, or attention-guided augmentation strategies, leading to strong performance under linear evaluation, fine-tuning, and semi-supervised protocols.

Masked modeling has also become a popular paradigm for skeleton representation learning. These methods mask partial joints, temporal segments, or motion patterns and train models to reconstruct or predict the missing skeleton information~\cite{wu2023skeletonmae,mao2023masked,wu2024macdiff,abdelfattah2024s}. By inferring invisible spatial-temporal structures from visible observations, masked modeling encourages the model to capture both local body dependencies and long-range motion dynamics. Recent masked skeleton modeling methods further design stronger reconstruction targets, motion-aware masking strategies, or predictive learning objectives, achieving competitive performance on standard skeleton benchmarks.

Despite their effectiveness, most existing self-supervised skeleton methods still rely on skeleton-specific backbones, graph structures, skeleton tokens, or carefully designed skeleton pretraining objectives. Their input representations are usually developed under fixed skeleton layouts with predefined joint numbers and joint orders. In contrast, our work investigates whether large-scale vision-pretrained masked models can be directly adapted to skeleton representation learning through a simple image-like interface. Rather than designing a new skeleton-specific SSL objective or architecture, we focus on bridging the representation gap between sparse skeleton sequences and dense image-pretrained models, enabling cross-format and universal skeleton representation learning across different skeleton layouts.

\subsection{Self-Supervised Representation Learning}
Self-supervised learning has gained significant traction in computer vision for its ability to learn effective representations without manual annotations. 
Contrastive learning methods exploit augmentation invariance to learn instance-discriminative features from images and videos~\cite{he2020momentum, chen2020simple, qian2021spatiotemporal}.
More recently, masked modeling has emerged as a powerful alternative to contrastive methods.
MAE~\cite{he2022masked} reconstructs masked pixel values using an asymmetric encoder-decoder design, providing a simple yet effective framework for visual representation learning. 
BEiT~\cite{bao2022beit} follows a mask-then-predict strategy, using visual tokens generated by a pre-trained tokenizer as prediction targets. 
%
MaskFeat~\cite{wei2022masked} instead predicts HOG features rather than raw pixels, providing a task-agnostic learning objective.
DiffMAE~\cite{wei2023diffusion} further enhances MAE by introducing a denoising diffusion process to iteratively reconstruct masked regions.
In this work, we evaluate our proposed S2I representation using both MAE and DiffMAE pretrained models.

\section{Method}
\subsection{Skeleton-to-Image Encoding}
\label{ref:S2I}
As discussed in Section~\ref{sec:intro}, our objective is to leverage vision-pretrained models for skeleton representation learning. This enables the effective utilization of large-scale vision models and their pretrained weights for skeleton tasks. Furthermore, adopting a unified vision model facilitates multi-modal action recognition, where diverse data modalities can be seamlessly integrated.

To achieve this, skeleton data must be reformatted into a representation compatible with image-based models. Specifically, it is essential to encode spatial-temporal information from skeleton sequences in a form analogous to image data, thereby enabling knowledge transfer from pretrained vision models.
However, a fundamental challenge arises from the inherent differences in data structures. While image data is typically represented as tensors of size $3 \times 224 \times 224$ for vision models, skeleton sequences are structured as $T \times J \times 3$, where $T$ denotes the temporal length of the sequence, and $J \times 3$ represents the 3D coordinates $(x, y, z)$ of skeleton joints.
To bridge this discrepancy, we propose a straightforward yet effective mapping strategy. The 3D joint coordinates $(x, y, z)$ are directly assigned to the RGB channels of an image, with each spatial axis corresponding to one color channel. The remaining challenge is to project the $T \times J$ spatial-temporal data into the $224 \times 224$ image, ensuring compatibility with standard vision model inputs.

\begin{figure*}[t]
    \centering
    \includegraphics[width=1\linewidth]{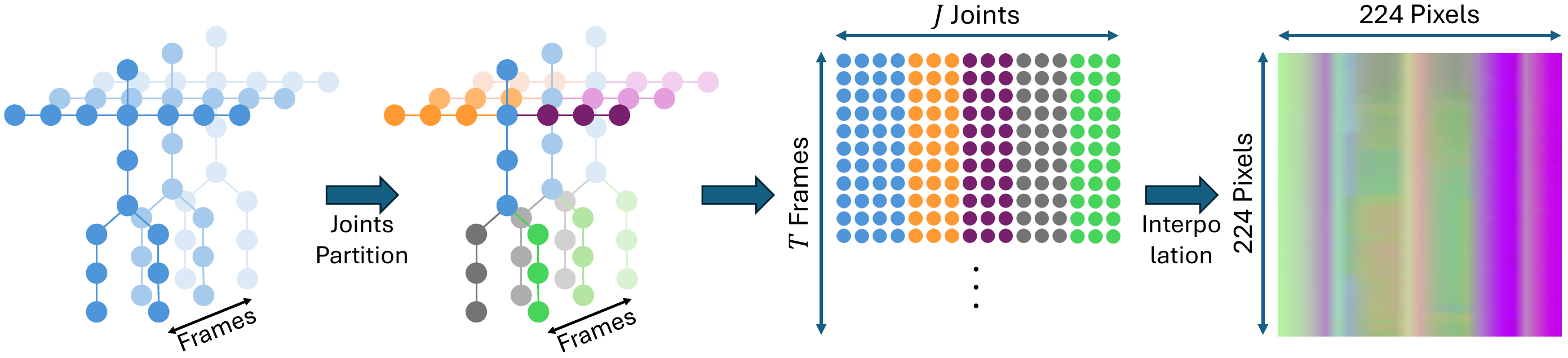}
    \caption{
     Illustration of the Skeleton-to-Image Encoding (S2I) process, which transforms skeleton sequences into image-like representations via joint partitioning, temporal stacking, and interpolation.
    }
    \label{fig:S2I}
\end{figure*}

To bridge the gap between skeleton sequences and image-based model inputs, we propose \textbf{Skeleton-to-Image Encoding (S2I)}, a novel representation that transforms skeleton sequences into dense, image-like data compatible with vision models (e.g., MAE, DiffMAE), as illustrated in Figure~\ref{fig:S2I}.
Specifically, we partition the human skeleton into five semantic body parts: spine, left arm, right arm, left leg, and right leg. This strategy, widely adopted in prior works~\cite{li2020dynamic, cai2019exploiting, zhang2021multi}, follows the kinematic structure of the human body and can be generalized across skeleton formats. 

Within each part, joints are arranged along their kinematic chain, ordered by distance from the torso. For example, joints in the left leg are sequenced as: left hip → left knee → left ankle → left foot. This ordering preserves the spatial relationships inherent in the skeleton structure.
For temporal modeling, the 3D positions of each joint across $T$ frames are stacked to form a spatio-temporal feature map of size $T \times J$, where $(x, y, z)$ coordinates are assigned to the RGB channels of the image.
Finally, to match the input requirements of vision models (e.g., $224 \times 224$), we apply linear interpolation along both temporal and joint dimensions, resizing them independently to a fixed resolution of $224 \times 224$. 
This yields an image-like representation of the skeleton sequence while preserving essential spatial-temporal patterns.


\vspace{-0.2cm}
\subsection{Vision Representation Models}
Our proposed S2I reformats skeleton data into image-like data, enabling seamless application of vision-pretrained models for skeleton representation learning. Unlike skeleton networks, our approach leverages the general architecture and pretrained weights of powerful vision models.

In this work, we evaluate two representative vision models to demonstrate the effectiveness of our representation: MAE~\cite{he2022masked} and DiffMAE~\cite{wei2023diffusion}. Both models are originally designed for image representation learning and pretrained on large-scale ImageNet, providing rich visual priors that can be effectively transferred to the skeleton domain through our unified representation.
\textbf{MAE} learns visual representations by reconstructing masked image patches from the visible context. Leveraging our S2I encoding to convert skeleton sequences into image-like data, 
we initialize MAE with ImageNet-pretrained weights and perform skeleton pretraining via masked reconstruction.
%
\textbf{DiffMAE} further enhances this framework by incorporating iterative denoising processes inspired by diffusion models.
Using the same S2I inputs, we similarly initialize DiffMAE from ImageNet-pretrained weights and conduct skeleton pretraining through diffusion-based reconstruction.
%

The skeleton-pretrained models are evaluated on downstream skeleton tasks to the effectiveness of adapting vision-pretrained models through the proposed S2I representation.
While we focus on MAE and DiffMAE in this work, our S2I representation is compatible with a broad range of vision models, including emerging generative and multimodal architectures.
%



\noindent\textbf{Application \& Advantage.}
Our S2I reformats skeleton sequences into image-like representations that are inherently robust to variations in skeleton structures. By abstracting skeleton data into a unified format, 
it enables seamless representation of diverse skeleton layouts without relying on dataset-specific joint definitions.
In contrast, conventional skeleton-based methods are tightly coupled to homogeneous skeleton formats, assuming fixed joint numbers, which limits their scalability and generalization across datasets with differing skeleton structures.
Even recent work~\cite{yang2021skeleton, liu2022collaborating} has attempted cross-format settings, but their evaluations still rely on shared joint subsets, thereby fundamentally adhering to homogeneous representations.

Our approach differs by introducing a format-agnostic representation paradigm. 
By decoupling skeleton representations from dataset-specific joint configurations,
it enables seamless integration of skeleton data from heterogeneous sources.
%
As a result, S2I naturally supports \textbf{cross-format representation learning}, allowing models trained on one skeleton format to generalize to others.
%
It also enables \textbf{universal representation pretraining} by jointly leveraging diverse skeleton datasets, similar to practices in large-scale image pretraining, without requiring task-specific architectural changes.
A visual comparison between conventional skeleton pipelines and the two proposed settings is provided in the Supplementary Material D.

%

\vspace{-0.15cm}
\subsection{Mask Sampling Strategy}
The effectiveness of masked modeling largely depends on the masking strategy employed during pretraining. To optimize representation learning for skeleton data, we investigate several masking strategies applicable to both image-based and skeleton-specific contexts.
\textbf{Random Masking} is the standard approach used in image-based masked modeling~\cite{he2022masked}, where image patches are randomly masked without considering spatial relationships. Formally, given a mask ratio $r$, we randomly select $\lfloor r \times N \rfloor$ patches to mask from the total $N$ patches in the skeleton image representation.
\textbf{Block Masking} increases pretraining task difficulty by masking contiguous regions of the input. Starting from randomly selected seed positions, blocks of adjacent patches are masked together, encouraging the model to learn stronger local structural relationships. 
Beyond these general strategies, we introduce two skeleton-specific masking schemes designed to better capture the unique spatial-temporal structure of human motion:
\textbf{Joint Masking} focuses on the spatial domain by randomly masking joints across the skeleton body, challenging the model to infer missing joint positions based on articulated body structure.
\textbf{Temporal Masking} targets the temporal dimension by masking entire frames or temporal slices, encouraging the model to capture dynamic motion patterns from partial sequences.

\vspace{-0.25cm}
\subsection{Training objectives}
We adopt a two-stage training pipeline for skeleton action recognition, leveraging our Skeleton-to-Image (S2I) representation to adapt vision-pretrained models to the skeleton domain.

In the first stage, we perform \textit{self-supervised skeleton representation learning} by applying masked modeling on our S2I representation, using the ImageNet-pretrained MAE~\cite{he2022masked} and DiffMAE~\cite{wei2023diffusion} as backbones.
%
Given a skeleton sequence $\mathbf{X} \in \mathbb{R}^{3 \times 224 \times 224}$ transformed via S2I, we apply masked modeling to learn skeleton representations:
%
%
For MAE, we minimize the reconstruction loss between the original input and the reconstruction of the masked patches:
\begin{equation}
\mathcal{L}_{\text{MAE}} = \frac{1}{|\mathcal{M}|} \sum_{i \in \mathcal{M}} \left\| \hat{\mathbf{X}}_i - \mathbf{X}_i \right\|_2^2,
\end{equation}
where $\mathcal{M}$ denotes the set of masked patches, $\hat{\mathbf{X}}_i$ is the reconstructed patch, and $\mathbf{X}_i$ is the ground truth.
For DiffMAE, we follow~\cite{wei2023diffusion} and reconstruct the masked regions through a denoising diffusion process, conditioned on the visible parts. The loss is defined as:
\begin{equation}
\mathcal{L}_{\text{DiffMAE}} = \mathbb{E}_{t, x_0, \epsilon} \left\| x_m^0 - D_\theta \left( x_m^t, t, E_\phi(x_v^0) \right) \right\|_2^2,
\end{equation}
where $x_m^0$ is the original masked region, $x_m^t$ is the noised version at diffusion step $t$, $x_v^0$ is the visible region, $E_\phi$ denotes the encoder, and $D_\theta$ is the diffusion decoder.


In the second stage, we evaluate the skeleton-pretrained encoders on downstream \textit{skeleton action recognition} tasks by attaching a classification head and optimizing with cross-entropy loss:
\begin{equation}
\mathcal{L}_{\text{CE}} = - \sum_{c=1}^{C} y_c \log \hat{y}_c,
\end{equation}
where $C$ is the number of classes, $y_c$ is the ground-truth label, and $\hat{y}_c$ is the predicted probability.
Depending on the evaluation protocol, we either perform linear probing with a frozen encoder or fine-tune the entire model.


\section{Experiments}

\begin{figure*}[t]
\begin{minipage}[t]{0.55\textwidth}
    \centering
    \vspace{-38mm}
    \resizebox{\linewidth}{!}{
    \begin{tabular}{lcc|cc}
    \toprule
     & Image Pretrain & Skeleton Pretrain & MAE & DiffMAE \\
    \midrule
    \multirow{4}{*}{Linear-Probe} 
      &   &  &   52.0 &  52.0 \\
      &   \checkmark &  & 72.2 & 71.3  \\
      &    &  \checkmark  & 76.3 & 78.6  \\
      &  \checkmark & \checkmark & 81.4 & \textbf{83.1}  \\
    \midrule
    \multirow{4}{*}{Fine-tune} 
      &  &  & 82.8 & 82.8 \\
      &   \checkmark &  & 86.8 & 86.5 \\
      &    &  \checkmark  & 88.1 &  88.6 \\
      &   \checkmark & \checkmark & 90.5 & \textbf{91.0}  \\
    \bottomrule
    \end{tabular}
    }
    \captionof{table}{Ablation study of image pretrain, skeleton pretrain on NTU-60 C-sub.}
    \label{tab:imageweight}
\end{minipage}
\hfill
\begin{minipage}[t]{0.42\textwidth}
    \centering
    \includegraphics[width=0.95\linewidth]{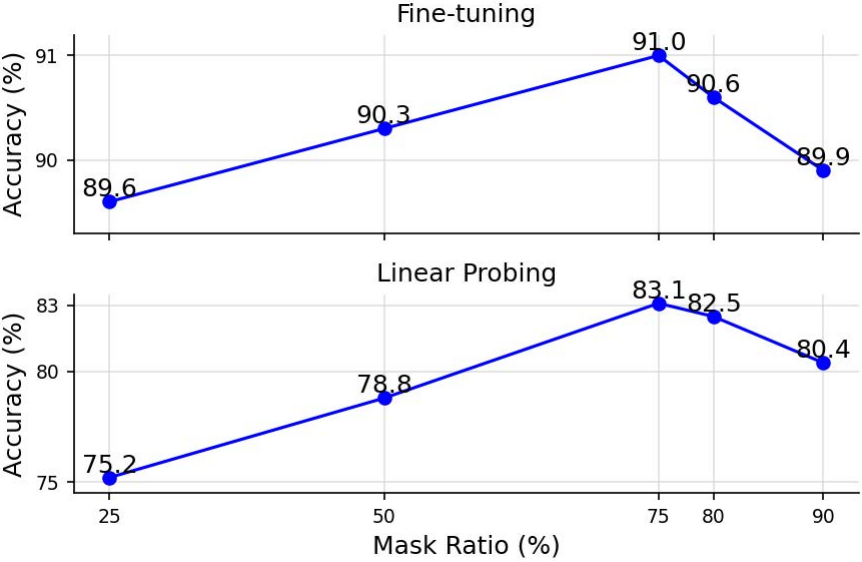}
    \captionof{figure}{Effect of mask ratio on NTU-60 C-sub.}
    \label{fig:mask_ratio_ablation}
\end{minipage}
\vspace{-0.3cm}
\end{figure*}

\begin{table*}[t]
    \centering
    \begin{minipage}[t]{0.48\linewidth}
        \centering
        \resizebox{\linewidth}{!}{
        \begin{tabular}{l|c|cc}
        \toprule
        Strategy & Ratio & Linear Probe & Fine-tune \\
        \midrule
        Joint & 75 & 82.3 & 90.4 \\
        Temporal & 75 & 82.9 & 90.7 \\
        Random & 75 & \textbf{83.1} & \textbf{91.0} \\
        Group & 75 & 81.3 & 90.3 \\
        \bottomrule
        \end{tabular}
        }
        \caption{Comparison of masking strategies on NTU-60 C-sub.}
        \label{tab:mask_strategy}
    \end{minipage}
    \hfill
    \begin{minipage}[t]{0.48\linewidth}
        \centering
        \resizebox{0.8\linewidth}{!}{
        \begin{tabular}{l|cc}
        \toprule
        Method & Linear Probe & Fine-tune \\
        \midrule
        Joint & 83.1 & 91.0 \\
        Motion & 70.5 & 89.7 \\
        Bone & 81.0 & 90.8 \\
        \midrule
        3-stream & \textbf{85.8} & \textbf{93.1} \\
        \bottomrule
        \end{tabular}
        }
        \caption{Comparison of skeleton modalities on NTU-60 C-sub.}
        \label{tab:skeleton_modalities}
    \end{minipage}
    \vspace{-0.6cm}
\end{table*}

\subsection{Datasets}
For evaluation, we conduct experiments on five datasets: NTU RGB+D 60 dataset (NTU-60)~\cite{Shahroudy_2016_CVPR}, NTU RGB+D 120 dataset
(NTU-120)~\cite{ntu120}, PKU-MMD ~\cite{pkummd}, Northwestern-UCLA (NW-UCLA)~\cite{wang2014cross}, and Toyota Smarthome (Toyota)~\cite{Das_2019_ICCV}. The first three use 25-joint skeletons, while NW-UCLA and Toyota contain 20 and 13 joints.

\noindent\textbf{NTU-60}~\cite{Shahroudy_2016_CVPR} is the most widely used benchmark for skeleton action recognition, comprising 56,880 skeleton sequences across 60 action classes performed by 40 subjects. Each sample contains at most two subjects, captured from three camera views using Kinect v2. In our experiments, we follow the standard cross-subject (C-sub) and cross-view (C-view) evaluation protocols.

\noindent\textbf{NTU-120}~\cite{ntu120} is the largest skeleton-based action recognition dataset to date, containing 114,480 samples across 120 action classes, collected from 106 subjects across 32 setups with varying locations and backgrounds. We follow the official cross-subject (C-sub) and cross-setup (C-set) evaluation protocols for benchmarking.

\noindent\textbf{PKU-MMD}~\cite{pkummd} is a large-scale benchmark for 3D human action understanding, featuring approximately 20,000 action instances in 51 categories. PKU-MMD consists of two subsets: Part I (easier, cleaner data) and Part II (challenging, with significant skeleton noise from large view variations). We follow the cross-subject protocol for both subsets in our experiments.

\noindent\textbf{NW-UCLA}~\cite{wang2014cross} contains 1,494 action samples from 10 classes, performed by 10 subjects and captured using three Kinect v1 cameras. Each skeleton consists of 20 joints. Following~\cite{wang2014cross}, we use samples from cameras V1 and V2 for training, and samples from camera V3 for testing.

\noindent\textbf{Toyota}~\cite{Das_2019_ICCV} is a real-world dataset for daily living activity recognition, containing 16,115 videos across 31 classes. 
We conduct transfer learning experiments under both cross-subject (CS) and cross-view1 (CV1) settings.

\subsection{Experimental Setup}
\label{sec: Experimental Setup}
\noindent\textbf{Network Architecture.} 
For both MAE and DiffMAE, we adopt the ViT-B architecture as the encoder. The decoder consists of eight transformer blocks, each with a channel dimension of 512. 
In DiffMAE, the decoder follows a cross-self attention design inspired by the original Transformer encoder-decoder structure. Specifically, in each decoder block, the noised tokens first attend to the visible latent representations via a cross-attention layer, followed by self-attention among noise tokens to refine predictions. 
~\footnote{Detailed network configurations and more implementation details are provided in Supplementary Material A, B, and C.}

\newcounter{archfootnote}
\setcounter{archfootnote}{\value{footnote}}

\noindent\textbf{Data Processing Details.} For NTU-60, NTU RGB-120, PKU-MMD dataset, and Toyota, we follow the data pre-processing in~\cite{zhang2020semantics}, sequence level translation based on the first frame is performed to be invariant
to the initial positions. If one frame contains two persons,
we split the frame into two frames by making each frame
contain one human skeleton. For the NW-UCLA dataset, we adopt the data pre-processing in~\cite{cheng2020skeleton}. 

\noindent\textbf{Implementation Details.}  
For the self-supervised skeleton learning stage, we initialize the models with ImageNet-pretrained MAE and DiffMAE checkpoints, and follow their default implementation configurations~\cite{he2022masked, wei2023diffusion}.~\footnotemark[\value{archfootnote}]  
For skeleton action recognition, we use SGD with Nesterov momentum (0.9) for linear probing, with an initial learning rate of 0.2 decayed via cosine annealing.  
For fine-tuning, we employ the AdamW optimizer with an initial learning rate of 0.001, decayed by cosine annealing.  
All skeleton action recognition experiments are trained for 100 epochs. The batch size is set to 128 for all datasets, except for NW-UCLA dataset, where a smaller batch size of 32 is used.
Our method is implemented in PyTorch, and all experiments are conducted on NVIDIA RTX A6000 GPUs.

\subsection{Ablation Study}
We conduct ablation studies on the NTU-60 C-sub to evaluate three key aspects of our method: the effectiveness of Skeleton-to-Image representation, the impact of image-pretrained weights, and the choice of masking strategies.


\begin{table*}[t]
\centering
\begin{tabular}{l|cc|cc|cc}
\toprule
\multirow{2}{*}{Method} & \multicolumn{2}{c|}{NTU-60} & \multicolumn{2}{c|}{NTU-120} & \multicolumn{2}{c}{PKU-MMD} \\
\cmidrule{2-7}
 & C-sub & C-view & C-sub & C-set & Phase I & Phase II \\
\midrule
LongT GAN~\cite{zheng2018unsupervised} & 39.1 & 48.1 & - & - & 67.7 & 26.0 \\
P\&C~\cite{su2020predict} & 50.7 & 76.3 & 42.7 & 41.7 & 59.9 & 25.5 \\
MS$^2$L~\cite{lin2020ms2l} & 52.6 & - & - & - & 64.9 & 27.6 \\
SkeletonMAE~\cite{wu2023skeletonmae} & 74.8 & 77.7 & 72.5 & 73.5 & 82.8 & 36.1 \\
3s-SkeletonCLR~\cite{li20213d} & 75.0 & 79.8 & 60.7 & 62.6 & 85.3 & - \\
3s-Colorization~\cite{yang2021skeleton} & 75.2 & 83.1 & - & - & - & - \\
ISC~\cite{thoker2021skeleton} & 76.3 & 85.2 & 67.1 & 67.9 & 80.9 & 36.0 \\
GL-Transformer~\cite{kim2022global} & 76.3 & 83.8 & 66.0 & 68.7 & - & - \\
3s-CrosSCLR~\cite{li20213d} & 77.8 & 83.4 & 67.9 & 66.7 & 84.9 & 21.2 \\
3s-AimCLR~\cite{guo2022contrastive} & 78.9 & 83.8 & 68.2 & 68.8 & 87.4 & 39.5 \\
CMD~\cite{mao2022cmd} & 79.4 & 86.9 & 70.3 & 71.5 & - & 43.0 \\
3s-CPM~\cite{zhang2022contrastive} & 83.2 & 87.0 & 73.0 & 74.0 & 90.7 & 51.5 \\
3s-ActCLR~\cite{lin2023actionlet} & 84.3 & 88.8 & 74.3 & 75.7 & - & - \\
MAMP~\cite{mao2023masked}  & 84.9 & 89.1 &78.6 & 79.1 & 92.2 & 53.8 \\
S-JEPA~\cite{abdelfattah2024s} & 85.3 & \underline{89.8} & \textbf{79.6} & 79.9 & 92.2 & 53.5 \\
MacDiff~\cite{wu2024macdiff} & \textbf{86.4} & \textbf{91.0} & \underline{79.4} & \underline{80.2} & \textbf{92.8} & - \\
\midrule
S2I (Ours) & 83.1 & 88.0 & 75.0 & 75.5 & 88.0 &  \underline{57.4} \\
3s-S2I (Ours) & \underline{85.8} & 89.7 & 78.9 & \textbf{80.3} & \underline{92.3} & \textbf{62.0} \\
\bottomrule
\end{tabular}
\caption{Comparison of Linear Evaluation results on NTU 60, NTU and PKU datasets. 3s- represents the ensemble results of joint(J), bone(B) and motion(M) streams. \textbf{Bold} and \underline{underlined} indicate the best and second best results, respectively. The same notation applies throughout.}
\vspace{-0.3cm}
\label{tab:comparison}
\end{table*}

\noindent\textbf{Effectiveness of Skeleton-to-Image Representation with Vision Models.}
To evaluate the viability of using image-based models for skeleton representation learning, we evaluate MAE and DiffMAE architectures in combination with our proposed Skeleton-to-Image Encoding. As shown in Table~\ref{tab:imageweight}, both models achieve strong performance after skeleton SSL pretraining, followed by linear probing or fine-tuning, demonstrating the viability of vision models for skeleton representation learning when equipped with appropriate representation strategies.
These results demonstrate that, despite being designed for dense image data, vision models can effectively process structured skeleton sequences when equipped with appropriate representations. Skeleton-to-Image Encoding serves as an effective bridge, enabling the reuse of powerful vision models without task-specific architectural modifications.

\noindent\textbf{Impact of Image-Pretrained Weights.}
We examine the benefit of ImageNet-pretrained weights by comparing models trained from scratch and models initialized with pretrained weights. Table~\ref{tab:imageweight} shows that image-pretraining yields substantial gains in both linear probing and fine-tuning settings.
%
%
In the linear probing scenario, pretrained MAE improves from 52.0\% (scratch) to 72.2\%, while pretrained DiffMAE improves from 52.0\% to 71.3\%. This substantial gap highlights the benefit of transferring generic visual representations to the skeleton domain. Furthermore, even after Skeleton Pretrain, pretraining with image data continues to provide notable gains, underscoring the importance of leveraging large-scale image-pretrained weights.
These results demonstrate that image-pretrained weights offer rich and transferable visual knowledge, serving as a strong initialization for skeleton representation learning and significantly enhancing performance.
Given DiffMAE's consistently superior performance, we adopt it as the \textbf{default backbone} for subsequent studies and comparisons.

\noindent\textbf{Effect of Masking Ratio.}
We investigate the influence of different masking ratios under the NTU-60 C-sub setting using DiffMAE. As shown in Figure~\ref{fig:mask_ratio_ablation}, increasing the masking ratio generally improves representation quality, with 75\% yielding the best results in both linear probing and fine-tuning.
Notably, linear probing performance shows a larger gain (+8\% improvement), while fine-tuning results are less sensitive (around 1\% difference). 
%
Based on these findings, we adopt a \textbf{75\%} masking ratio as the default setting for subsequent ablation studies and main comparisons.

\noindent
\textbf{Comparison of Masking Strategies.}
We compare various masking strategies: joint, temporal, random, and group masking. Results in Table~\ref{tab:mask_strategy} indicate that random masking consistently outperforms others.
While joint and temporal masking provide competitive results, group masking underperforms, suggesting that overly structured masking limits the model's capacity to capture diverse patterns. Consequently, we adopt \textbf{random masking with 75\% ratio} as our default strategy.





\begin{table*}[t]
\centering
\begin{minipage}[t]{0.53\textwidth}
    \centering
    \resizebox{\linewidth}{!}{
    \begin{tabular}{l|cc|cc}
    \toprule
    \multirow{2}{*}{Method} & \multicolumn{2}{c|}{NTU-60} & \multicolumn{2}{c}{NTU-120} \\
    \cmidrule{2-5}
     & C-sub & C-view & C-sub & C-set \\
    \midrule
    AimCLR (STTFormer)~\cite{guo2022contrastive} & 83.9 & 90.4 & 74.6 & 77.2 \\
    CrosSCLR (STTFormer)~\cite{li20213d} & 84.6 & 90.5 & 75.0 & 77.9 \\
    CPM~\cite{zhang2022contrastive} & 84.8 & 91.1 & 78.4 & 78.9 \\
    3s-CrosSCLR~\cite{li20213d} & 86.2 & 92.5 & 80.5 & 80.4 \\
    3s-AimCLR~\cite{guo2022contrastive} & 86.9 & 92.8 & 80.1 & 80.9 \\
    SkeletonMAE~\cite{wu2023skeletonmae} & 86.6 & 92.9 & 76.8 & 79.1 \\
    3s-Colorization~\cite{yang2021skeleton} & 88.0 & 94.9 & - & - \\
    3s-ActCLR~\cite{lin2023actionlet} & 88.2 & 93.9 & 82.1 & 84.6 \\
    MCC~\cite{su2021self} & 89.7 & 96.3 & 81.3 & 83.5 \\
    ViA~\cite{yang2024view} & 89.6 & 96.4 & 85.0 & 86.5 \\
    3s-Hi-TRS~\cite{chen2022hierarchically} & 90.0 & 95.7 & 85.3 & 87.4 \\
    MAMP~\cite{mao2023masked} & \textbf{93.1} & 97.5 & 90.0 & \textbf{91.3} \\
    S-JEPA~\cite{abdelfattah2024s} & \textbf{93.1} & \underline{97.6} & \textbf{90.3} & \textbf{91.3} \\ 
    MacDiff~\cite{wu2024macdiff} & 92.7 & 97.3 & - & - \\ 
    \midrule
    S2I (Ours) & 91.0 & 96.7 & 86.6 & 87.9 \\
    3s-S2I (Ours) & \textbf{93.1} & \textbf{97.7} & \underline{90.2} & \underline{91.2} \\
    \bottomrule
    \end{tabular}
    }
    \caption{Comparison of fine-tuned results on the NTU-60 and NTU-120 datasets.}
    \label{tab:finetune_comparison}
\end{minipage}
\hfill
\begin{minipage}[t]{0.41\textwidth}
    \centering
    \resizebox{\linewidth}{!}{
    \begin{tabular}{l|cc|cc}
    \toprule
    \multirow{3}{*}{Method} & \multicolumn{4}{c}{NTU-60} \\
    \cmidrule{2-5}
    & \multicolumn{2}{c|}{C-sub} & \multicolumn{2}{c}{C-view} \\
    \cmidrule{2-5}
    & (1\%) & (10\%) & (1\%) & (10\%) \\
    \midrule
    LongT GAN~\cite{zheng2018unsupervised} & 35.2 & 62.0 & - & - \\
    MS$^2$L~\cite{lin2020ms2l} & 33.1 & 65.1 & - & - \\
    ASSL~\cite{si2020adversarial} & - & 64.3 & - & 69.8 \\
    ISC~\cite{thoker2021skeleton} & 35.7 & 65.9 & 38.1 & 72.5 \\
    3s-CrosSCLR~\cite{li20213d} & 51.1 & 74.4 & 50.0 & 77.8 \\
    3s-Colorization~\cite{yang2021skeleton} & 48.3 & 71.7 & 52.5 & 78.9 \\
    CMD~\cite{mao2022cmd} & 50.6 & 75.4 & 53.0 & 80.2 \\
    3s-Hi-TRS~\cite{chen2022hierarchically} & 49.3 & 77.7 & 51.5 & 81.1 \\
    3s-AimCLR~\cite{guo2022contrastive} & 54.8 & 78.2 & 54.3 & 81.6 \\
    3s-CMD~\cite{mao2022cmd} & 55.6 & 79.0 & 55.5 & 82.4 \\
    CPM~\cite{zhang2022contrastive} & 56.7 & 73.0 & 57.5 & 77.1 \\
    SkeletonMAE~\cite{wu2023skeletonmae} & 54.4 & 80.6 & 54.6 & 83.5 \\
    MAMP~\cite{mao2023masked} & 66.0 & 88.0 & 68.7 & 91.5 \\
    S-JEPA~\cite{abdelfattah2024s} & 67.5 & \textbf{88.4} & 69.1 & 91.4 \\ 
    MacDiff~\cite{wu2024macdiff}  & 65.6 & \underline{88.2} & \underline{77.3} & \textbf{92.5} \\
    \midrule
    S2I (Ours) & \underline{71.4} & 84.8 & 73.3 & 87.8\\
    3s-S2I (Ours) & \textbf{75.2} & \underline{88.3} & \textbf{77.7} & \underline{91.7}\\
    \bottomrule
    \end{tabular}
    }
    \caption{Comparison of semi-supervised results on the NTU-60 dataset.}
    \label{tab:semi_supervised}
\end{minipage}
\vspace{-0.6cm}
\end{table*}

\begin{table*}[t]
\begin{minipage}[t]{0.45\textwidth}
    \centering
    \resizebox{1\linewidth}{!}{
    \begin{tabular}{l|ccc}
    \toprule
    \multirow{2}{*}{Method} & \multicolumn{3}{c}{To PKU-II} \\
    \cmidrule{2-4}
     & NTU-60 & NTU-120 & PKU-I \\
    \midrule
    LongT GAN~\cite{zheng2018unsupervised} & 44.8 & - & 43.6 \\
    MS$^2$L~\cite{lin2020ms2l} & 45.8 & - & 44.1 \\
    ISC~\cite{thoker2021skeleton} & 51.1 & 52.3 & 45.1 \\
    CMD~\cite{mao2022cmd}  & 56.0 & 57.0 & - \\
    SkeletonMAE~\cite{wu2023skeletonmae} & 58.4 & 61.0 & 62.5 \\
    MAMP~\cite{mao2023masked}  & 70.6 & 73.2 & 70.1 \\
    S-JEPA~\cite{abdelfattah2024s} & 71.4 & \textbf{74.2} & 70.9 \\ 
    MacDiff~\cite{wu2024macdiff} & \underline{72.2} & {73.4} & \underline{71.4} \\
    \midrule
    S2I (ours) & 70.2 & 71.6 & 68.1 \\
    3s-S2I (ours) & \textbf{72.6} & \underline{73.9} & \textbf{72.9} \\
    \bottomrule
    \end{tabular}
    }
    \caption{Comparison of transfer learning results on PKUMMD II dataset.}
    \label{tab:pku2_transfer}
\end{minipage}
\hfill
\begin{minipage}[t]{0.45\textwidth}
    \centering
    \resizebox{1\linewidth}{!}{
    \begin{tabular}{l|cc|c}
        \toprule
        \multirow{3}{*}{Method} & \multicolumn{3}{c}{NTU-60 (25 joints)} \\
        \cmidrule(lr){2-4}
         & \multicolumn{2}{c|}{Toyota (13 joints)} & NW-UCLA  \\
        & CS & CV1 & (20 joints) \\
        \midrule
        Colorization~\cite{yang2021skeleton} & - & - & \underline{93.3} \\
        UNIK~\cite{yang2021unik} & 63.1 & 22.9 & - \\
        ViA~\cite{yang2024view}  & 64.5 & 36.1 & - \\
        \midrule
        S2I (Ours) & \underline{65.4} & \underline{43.1} & 93.2 \\
        3s-S2I (Ours) & \textbf{70.1} & \textbf{53.8} & \textbf{94.2} \\
        \bottomrule
    \end{tabular}
    }
    \caption{Cross-format transfer learning results from NTU-60 to Toyota and NW-UCLA datasets.}
    \label{tab:cross_format}
\end{minipage}
\vspace{-0.2cm}
\end{table*}

\noindent
\textbf{More Skeleton Modalities.}
In skeleton-based recognition, joint data can be further augmented by deriving motion and bone modalities. We evaluate these modalities individually and in combination using DiffMAE fine-tuning to assess their effectiveness within vision models.
%
As shown in Table~\ref{tab:skeleton_modalities}, joint features perform strongly, while motion and bone streams offer complementary cues.
Fusing all three modalities significantly boosts performance, reaching 85.8\% (linear probing) and 93.1\% (fine-tuning). This confirms that this skeleton-modality fusion remains effective even when skeleton data is reformatted as images and processed by vision models. 
%
In subsequent main results, we report both the Joint stream (\textbf{S2I}) and the 3-stream fusion (\textbf{3s-S2I}).


\begin{table*}[t]
\centering
\begin{tabular}{l|cccccc}
\toprule
Method & NTU120-Csub & NTU60-Csub & PKU-I-CS & PKU-II-CS & Toyota-CS &  NW-UCLA \\
\midrule
Self-pretrain       & 86.6 & 91.0 & 94.0 & 65.8  & 64.5 & 91.6 \\
Universal-pretrain  & \textbf{87.0} & \textbf{91.6} & \textbf{95.2} & \textbf{71.1} & \textbf{68.0} & \textbf{93.5} \\
\bottomrule
\end{tabular}
\caption{Universal pretraining evaluation on multiple skeleton datasets.}
\vspace{-0.5cm}
\label{tab:universal_pretrain}
\end{table*}

\subsection{Comparison with the State-of-the-art Methods}

\noindent\textbf{Linear Evaluation Results.}
In linear evaluation, we freeze the pretrained backbone and train a supervised linear classifier on top.
Table~\ref{tab:comparison} summarizes results on NTU-60, NTU-120, and PKU-MMD. Despite using image-pretrained vision models without skeleton-specific architectures, our S2I achieves competitive performance compared to recent specialized methods.
Specifically, S2I attains 83.1\% and 88.0\% on NTU-60 (C-sub and C-view), and 75.0\% on NTU-120 (C-sub), demonstrating its effectiveness in bridging skeleton data with vision models.
By integrating joint, motion, and bone streams, 3s-S2I further enhances representation quality, achieving state-of-the-art results on NTU-120 C-set (80.3\%) and PKU II (62.0\%).
 
\noindent\textbf{Fine-tuned Evaluation Results.}
In a fine-tuned protocol, an MLP head is attached to the pre-trained backbone, and the whole network is fully fine-tuned. As shown in Table~\ref{tab:finetune_comparison}, our S2I achieves competitive performance on NTU-60 and NTU-120 without introducing any skeleton-specific architectural designs.
By bridging skeleton data with vision-pretrained models, S2I adapts effectively to skeleton tasks.
With multi-stream fusion, the 3-stream variant (3s-S2I) provides additional gains, further confirming that the proposed interface can be combined with standard skeleton modalities.

\noindent\textbf{Semi-supervised Evaluation Results.}
In the semi-supervised protocol, the classification head and the pretrained encoder are fine-tuned using only a small fraction of labeled training samples. 
We evaluate on NTU-60 with 1\% and 10\% labeled samples. 
As shown in Table~\ref{tab:semi_supervised}, S2I shows particularly strong performance in the extremely low-label regime. With only the joint stream, S2I achieves 71.4\% under the 1\% C-sub setting, outperforming all compared methods, including recent masked modeling and predictive learning approaches. Under the 1\% C-view setting, S2I also achieves competitive performance with 73.3\%, demonstrating the effectiveness of adapting vision-pretrained models through the proposed image-like interface when annotations are scarce. 
With multi-stream fusion, 3s-S2I further improves the results to 75.2\% and 77.7\% under the 1\% C-sub and C-view settings, respectively. These results suggest that S2I provides strong transferable representations in low-label regimes, while the 10\% setting remains competitive with recent skeleton-specific methods.

\noindent\textbf{Transfer Learning Evaluation Results.}
In the transfer learning protocol, models are pre-trained on a source dataset and then fine-tuned on a distinct target dataset, thereby assessing the generalization ability of the learned representations. 
In this setting, PKU-MMD II serves as the target dataset, while NTU-60, NTU-120, and PKU-MMD I are used as source datasets. 
As reported in Table~\ref{tab:pku2_transfer}, S2I achieves competitive transfer performance on PKU-MMD II across different source datasets. The 3-stream variant further improves transfer accuracy, suggesting that the proposed interface can benefit from complementary skeleton modalities. 
These results indicate that vision-pretrained skeleton representations learned through S2I generalize strongly across datasets. In particular, the 3-stream S2I variant achieves the best transfer performance when transferring from NTU-60 and PKU-MMD I to PKU-MMD II, demonstrating that the proposed representation is especially effective when the target domain differs from the source domain.

\subsection{Broader Applications}
As discussed in Section~\ref{ref:S2I}, S2I provides a unified representation that supports both Cross-Format Transfer Learning and Universal Skeleton Representation Learning. To highlight its advantages, we conduct experiments under these two settings.

\noindent\textbf{Cross-Format Transfer Learning Evaluation Results.}
We evaluate S2I on three heterogeneous skeleton datasets: NTU-60 (25 joints), Toyota (13 joints), and NW-UCLA (20 joints). Existing methods often rely on joint downsampling or interpolation to match skeleton formats, which can introduce information loss or noise.
In contrast, our S2I preserves structural information by converting skeleton sequences into a format-agnostic image representation. As shown in Table~\ref{tab:cross_format}, S2I achieves clear improvements over prior works.
Notably, 3s-S2I reaches 53.8\% on Toyota (CV1), surpassing existing methods by a significant margin. 

\noindent\textbf{Universal Representation Pretraining Evaluation Results.}
To assess the generalizability of our representation, we perform universal representation learning by aggregating training data from multiple datasets (NTU120-Csub, PKU-I-CS, PKU-II-CS, Toyota-CS, and NW-UCLA).~\footnote{More details of implementations and skeleton partitioning are provided in Supplementary Material E.} 
All experiments are conducted on joint data.
As shown in Table~\ref{tab:universal_pretrain}, universal pretraining consistently boosts performance on all evaluated datasets compared to individual dataset self-pretraining. 
Notably, we obtain substantial gains on PKUV2-CS (+5.3\%) and Toyota-CS (+3.5\%).

Collectively, the results from cross-format transfer learning and universal representation pretraining show that the main advantage of S2I emerges in transfer-oriented scenarios. By converting heterogeneous skeleton formats into a shared image-like representation, S2I can exploit substantially broader pretraining data than methods tied to fixed skeleton layouts, leading to more robust transfer to new datasets and formats.


\section{Conclusion}
\label{sec: conclusion}
In this paper, we propose Skeleton-to-Image Encoding (S2I), a simple yet effective representation that bridges spatio-temporal skeleton sequences with vision-pretrained models. By reformatting skeleton data into image-like structures, S2I enables the direct adoption of powerful pretrained vision models without requiring skeleton-specific architectural modifications.
Through extensive experiments on five benchmark datasets, we demonstrated that S2I achieves competitive performance for skeleton representation learning. 
More importantly, S2I shows stronger advantages in transfer-oriented scenarios, including cross-format transfer learning across heterogeneous skeleton layouts and universal skeleton pretraining over diverse datasets, where its unified image-like representation enables more robust generalization to new datasets and skeleton formats
%
In future work, we will explore extending S2I to larger vision and vision-language models, as well as its integration with RGB videos and other sensor modalities for multi-modal action recognition.

\section{Data Availability Statement}
The authors declare that the datasets used in this
paper are available at the following links:
\begin{itemize}
    \item NTU RGB+D~\cite{Shahroudy_2016_CVPR} and NTU RGB+D 120~\cite{ntu120}: \url{https://rose1.ntu.edu.sg/dataset/actionRecognition/}
    \item PKU-MMD~\cite{pkummd}: \url{https://struct002.github.io/PKUMMD/}
    \item Northwestern-UCLA ~\cite{wang2014cross}: \url{https://wangjiangb.github.io/my_data.html}
    \item Toyota Smarthome~\cite{Das_2019_ICCV}: \url{https://project.inria.fr/toyotasmarthome/}
    
\end{itemize}


\section{Declarations}
\noindent\textbf{Conflict of interest.}The authors have no competing interests to declare that are relevant to the content of this article.
\backmatter

\section*{Acknowledgements}
We gratefully acknowledge access to the Berzelius computing resource at the National Supercomputer
Centre, provided by the Knut and Alice Wallenberg Foundation. This work was supported by the
Wallenberg AI, Autonomous Systems and Software Program (WASP).

\bibliography{sn-bibliography}

\end{document}


\title[Article Title]{Supplementary Material for Skeleton-to-Image Encoding: Enabling Skeleton Representation Learning via Vision-Pretrained Models}

\author[1]{\fnm{Siyuan} \sur{Yang}}\email{siyuany@kth.se}

\author*[2]{\fnm{Jun} \sur{Liu}}\email{j.liu81@lancaster.ac.uk}
\author[3]{\fnm{Hao} \sur{Cheng}}\email{chenghao@hebut.edu.cn}
\author[4]{\fnm{Chong} \sur{Wang}}\email{wang1711@ntu.edu.sg}
\author[4]{\fnm{Shijian} \sur{Lu}}\email{Shijian.Lu@ntu.edu.sg}
\author[1]{\fnm{Hedvig} \sur{Kjellstrom}}\email{hedvig@kth.se}
\author[4]{\fnm{Weisi} \sur{Lin}}\email{wslin@ntu.edu.sg}
\author[4,5,6]{\fnm{Alex} \sur{Kot}}\email{eackot@ntu.edu.sg}

\affil[1]{\orgname{KTH Royal Institute of Technology},  \orgaddress{\city{Stockholm}, \country{Sweden}}}
\affil[2]{\orgname{Lancaster University}, \orgaddress{\city{Lancaster}, \country{UK}}}
\affil[3]{\orgname{Hebei University of Technology},  \orgaddress{\city{Tianjin}, \country{China}}}
\affil[4]{\orgname{Nanyang Technological University},  \orgaddress{\city{Singapore}, \country{Singapore}}}
\affil[5]{\orgname{Shenzhen MSU-BIT University},  \orgaddress{\city{Shenzhen}, \country{China}}}
\affil[6]{\orgname{VinUniversity},  \orgaddress{\city{Hanoi}, \country{Vietnam}}}

\maketitle

\section{Network Architecture and Configurations}
\label{sec: network configuration}
\noindent
\textbf{MAE~\cite{he2022masked}.} 
Our MAE implementation follows the asymmetric encoder-decoder architecture proposed in~\cite{he2022masked}. The encoder is a Vision Transformer (ViT-B/16)~\cite{dosovitskiy2021an}, where ``16'' denotes the patch size. 
It operates only on the visible (unmasked) patches, without using any mask tokens. Each patch is linearly projected and added with fixed sinusoidal positional embeddings.  
The encoder consists of 12 Transformer blocks, each composed of a multi-head self-attention layer~\cite{vaswani2017attention}, a feed-forward MLP, and LayerNorm~\cite{ba2016layer} applied before both submodules. A standard class token is used during fine-tuning but omitted during pretraining.
The decoder is lightweight and receives both the encoded visible patches and learnable mask tokens. It comprises 8 Transformer blocks~\cite{vaswani2017attention} with a hidden size of 512 and projects the output back to pixel space to reconstruct the original input.  
Reconstruction is performed only on masked patches, and the loss is computed as the mean squared error (MSE) between the predicted and ground-truth pixel values.

\noindent
\textbf{DiffMAE~\cite{wei2023diffusion}.}
Our implementation of DiffMAE follows the asymmetric encoder-decoder architecture proposed in~\cite{wei2023diffusion}, using ViT-B/16 as the encoder. The encoder processes only visible patches, and the decoder reconstructs masked regions from noisy inputs sampled through a forward diffusion process.
Following the original design, we append a LayerNorm to the encoder output, followed by a linear projection to align feature dimensions with the decoder. Fixed sinusoidal positional embeddings are added to both the encoder and decoder inputs. We do not use relative positional encodings or layer scale. Additionally, separate linear projections are applied to clean and noisy tokens, respectively.

Unlike MAE, which directly regresses pixel values, DiffMAE formulates masked region prediction as a conditional generation task via a diffusion-based denoising process.  
We adopt the \emph{cross decoder} variant, where each noisy token (corresponding to a masked patch) independently attends to encoder outputs via cross-attention, without interacting with other noise tokens. This avoids shortcut paths and promotes more effective encoder pretraining.
For the diffusion process, we follow a linear variance schedule~\cite{ho2020denoising} and set the number of timesteps to $T=1000$.

\section{Pretraining Setup and Strategies}
\label{sec: pretraining details}
\noindent
\textbf{MAE~\cite{he2022masked}.}
We follow the general pretraining setup of MAE~\cite{he2022masked}, with modifications to accommodate the smaller scale and different nature of skeleton datasets compared to ImageNet. Specifically, we use a batch size of 512 and train for 800 epochs.
For data augmentation, we employ a customized pipeline designed for S2I skeleton representation. In contrast to the original MAE, which uses standard image augmentations such as \texttt{ColorJitter} and \texttt{DropPath}, we adopt lightweight and semantically consistent transformations: random horizontal flip ($p=0.5$), random rotation, random affine scaling and translation, and additive Gaussian noise with standard deviation. All augmentations are applied before normalization.
All other components, including optimizer settings and learning rate schedule, remain consistent with the original MAE. The full configuration is summarized in Table~\ref{tab:pretrain}.


\begin{table}[t]
\centering
\caption{Pretraining settings for MAE on S2I representation.}
\label{tab:pretrain}
\small
\begin{tabular}{l|p{0.45\linewidth}}
\toprule
\textbf{Config} & \textbf{Value} \\
\midrule
Optimizer & AdamW \\
Base learning rate & $1.5 \times 10^{-4}$ \\
Weight decay & 0.05 \\
Optimizer momentum & $\beta_1, \beta_2 = 0.9, 0.95$ \\
Batch size & 512 \\
Learning rate schedule & cosine decay \\
Epochs & 800 \\
Warmup epochs & 40 \\
Augmentation & Horizontal flip $(p=0.5)$, Random rotation, Random affine, Gaussian noise \\
\bottomrule
\end{tabular}
\end{table}

\noindent
\textbf{DiffMAE~\cite{wei2023diffusion}.}
We follow the official DiffMAE pretraining setup on ImageNet, using the same optimizer, learning rate schedule, and number of training epochs (1600). To account for the smaller scale of skeleton dataset, we reduce the batch size to 512.
In place of \texttt{RandomResizedCrop}, we adopt a skeleton-related augmentation pipeline consisting of horizontal flip, random rotation, affine transformation, and additive Gaussian noise, as detailed above. The full pretraining configuration is summarized in Table~\ref{tab:diffmae_pretrain}.

For the diffusion process, we follow the noise schedule formulation in~\cite{wei2023diffusion}, where each forward sample $x_t^m$ is reparameterized as:
\[
x_t^m = \sqrt{\bar{\alpha}_t} x_0^m + \sqrt{1 - \bar{\alpha}_t} \epsilon,
\]
with $\alpha_t = 1 - \beta_t$ and $\bar{\alpha}_t = \prod_{i=1}^{t} \alpha_i$.
To control the noise level, we introduce a hyperparameter $\rho$ following~\cite{wei2023diffusion}, which modulates the variance schedule by exponentiating each $\beta_t$ as $\beta_t^\rho$. We set $\rho = 1.0$ by default, which recovers the standard linear schedule in~\cite{ho2020denoising}, where $\beta_t$ increases linearly from $10^{-4}$ to $0.02$.


\begin{table}[t]
\centering
\caption{Pretraining settings for DiffMAE on S2I representation.}
\label{tab:diffmae_pretrain}
\small
\begin{tabular}{l|p{0.45\linewidth}}
\toprule
\textbf{Config} & \textbf{Value} \\
\midrule
Optimizer & AdamW \\
Weight decay & 0.05 \\
Base learning rate & $1.5 \times 10^{-4}$ \\
Optimizer momentum & $\beta_1, \beta_2 = 0.9, 0.95$ \\
Batch size & 512 \\
Learning rate schedule & cosine decay \\
Epochs & 1600 \\
Warmup epochs & 40 \\
Augmentation & Horizontal flip $(p=0.5)$, Random rotation, Random affine, Gaussian noise \\
\bottomrule
\end{tabular}
\end{table}

\section{Downstream Training Protocols: Fine-tuning and Linear Probing}
\label{sec: implementation}
We extract features from the encoder output for downstream tasks, including fine-tuning and linear probing. Following the standard ViT design~\cite{dosovitskiy2021an}, which includes a class token, we append an auxiliary dummy token to the input sequence during pretraining, as in~\cite{he2022masked}. 
This token is treated as the class token and used for classification in both fine-tuning and linear probing.

\noindent
\textbf{Fine-tuning.}
Our fine-tuning protocol follows standard supervised ViT training~\cite{dosovitskiy2021an}, with configurations tailored for skeleton-based inputs. Specifically, we omit image-based data augmentation and regularization techniques such as RandAugment, label smoothing, mixup, and cutmix. Instead, we apply a skeleton-related augmentation pipeline, including horizontal flip, rotation, affine transformation, and Gaussian noise.
We use the AdamW optimizer with a base learning rate of $1\text{e}{-3}$ and a cosine learning rate decay schedule. Layer-wise learning rate decay is applied with a decay rate of 0.75, following~\cite{bao2022beit}. We set the batch size to 128 for NTU-60, NTU-120, PKU-MMD, and Toyota datasets, and reduce it to 32 for the smaller NW-UCLA dataset to prevent overfitting. The warmup period is fixed to 10 epochs, and training is performed for 100 epochs in total.

\begin{table}[t]
\centering
\caption{Fine-tuning settings for S2I representation.}
\label{tab:finetune}
\small
\begin{tabular}{l|p{0.45\linewidth}}
\toprule
\textbf{Config} & \textbf{Value} \\
\midrule
Optimizer & AdamW \\
Base learning rate & $1 \times 10^{-3}$ \\
Weight decay & 0.05 \\
Layer-wise lr decay & 0.75 \\
Batch size & 128 (NTU/PKU/Toyota), 32 (N-UCLA) \\
Learning rate schedule & cosine decay \\
Warmup epochs & 10 \\
Training epochs & 100 \\
Augmentation & Horizontal flip $(p=0.5)$, Random rotation, Random affine, Gaussian noise \\
\bottomrule
\end{tabular}
\end{table}

\noindent
\textbf{Linear probing.}
For linear probing, we freeze the encoder and train only a linear classification head on top of the extracted features. We follow a simplified setup without additional regularization such as weight decay, label smoothing, or mixup. 
We use SGD with a base learning rate of 0.2, momentum of 0.9, and a cosine learning-rate schedule. Training runs for 100 epochs with a 10-epoch warmup.
We use the same skeleton-related data augmentation as in fine-tuning. Detailed configurations are listed in Table~\ref{tab:linprobe}.

\section{Pipeline Comparison under Cross-Format and Universal Settings}
\label{sec: pipeline comparison}
\begin{figure*}[t]
    \centering
    \includegraphics[width=1\linewidth]{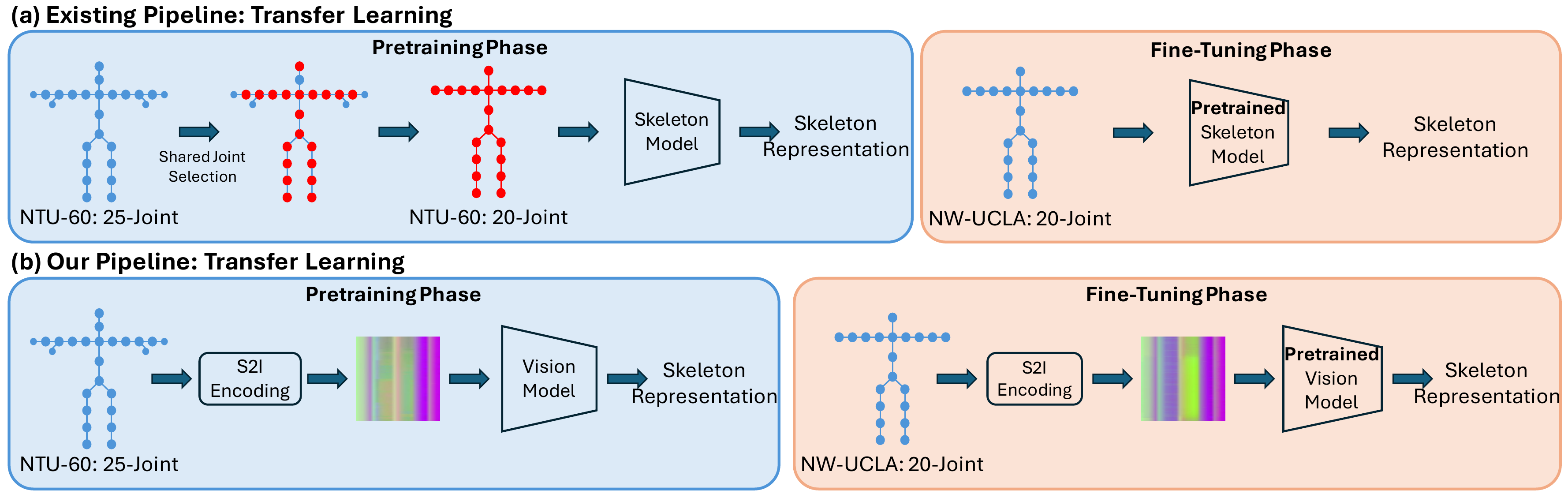}
    \caption{Comparison between the existing skeleton-specific pipeline and our proposed S2I-based pipeline for transfer learning across datasets. (a) Existing methods first align joint formats before pretraining and fine-tuning on target datasets, which may lead to information loss. (b) Our approach bypasses manual joint selection by encoding the raw skeleton sequence into an image via S2I, enabling unified processing across datasets using a vision model.}
    \label{fig:cross_format_pipeline_compare}
    \vspace{-0.3cm}
\end{figure*}

\noindent
\textbf{Cross-Format Representation Learning.}
To evaluate cross-format transfer performance, we compare our pipeline against the conventional transfer learning paradigm, using the transfer from NTU-60 to NW-UCLA as an example.
As illustrated in Figure~\ref{fig:cross_format_pipeline_compare}, existing approaches typically begin by manually downsampling the NTU-60 skeleton data (25 joints) to a shared 20-joint subset. A model is then pretrained on this reduced-format data and fine-tuned on NW-UCLA, which naturally adopts the 20-joint format. This process requires explicit joint alignment and may introduce structural mismatches, potentially degrading the quality of learned representations.

In contrast, our S2I-based pipeline removes the need for manual joint selection. We directly encode the raw skeleton sequences—regardless of their joint format—into semantically structured image-like representations. Both NTU-60 and NW-UCLA inputs are processed uniformly by the same visual backbone without any format-specific adaptation. This design enables seamless cross-format transfer while preserving the structural integrity of the original data.
Figure~\ref{fig:cross_format_pipeline_compare} provides a visual comparison of the two pipelines, highlighting the simplicity and universality of our approach for cross-format transfer learning.

\begin{table}[t]
\centering
\caption{Linear probing settings for S2I representation.}
\label{tab:linprobe}
\small
\begin{tabular}{l|p{0.45\linewidth}}
\toprule
\textbf{Config} & \textbf{Value} \\
\midrule
Optimizer & SGD \\
Base learning rate & $2 \times 10^{-1}$ \\
Weight decay & 0 \\
Optimizer momentum & 0.9 \\
Batch size & 128 \\
Learning rate schedule & cosine decay \\
Warmup epochs & 10 \\
Training epochs & 100 \\
Augmentation & Horizontal flip $(p=0.5)$, Random rotation, Random affine, Gaussian noise \\
\bottomrule
\end{tabular}
\vspace{-0.2cm}
\end{table}


\begin{figure*}[t]
    \centering
    \includegraphics[width=1\linewidth]{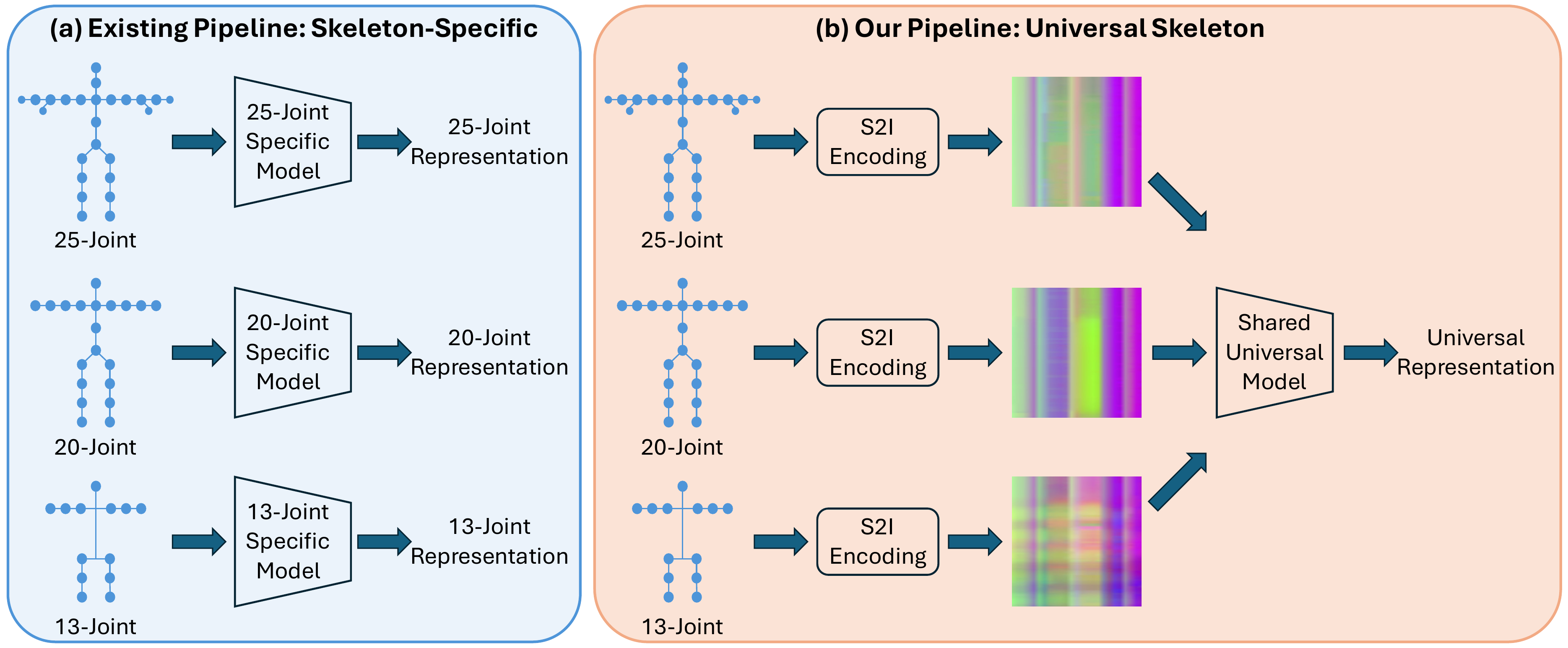}
    \caption{
    Comparison of skeleton-specific and universal representation learning pipelines. (a) Conventional methods require format-specific models for each skeleton format. (b) Our method encodes arbitrary skeleton formats into image representations via S2I, enabling unified pretraining with a single backbone model.}
    \label{fig:univerisal_pipeline_compare}
\end{figure*}

\noindent
\textbf{Universal Representation Pretraining.}
To facilitate universal skeleton representation learning across diverse datasets, we compare our unified pipeline with traditional skeleton-specific approaches. As shown in Figure~\ref{fig:univerisal_pipeline_compare}, conventional methods require designing and training separate models tailored to each joint format (e.g., 25-joint, 20-joint, 13-joint), resulting in limited scalability.

In contrast, our approach leverages the S2I (Skeleton-to-Image) encoding to transform skeleton sequences with arbitrary joint formats into structured image representations. These representations are then processed by a shared backbone model, enabling consistent and format-agnostic feature learning.
This unified design eliminates the need for joint-level alignment or model reconfiguration, allowing all skeleton datasets—regardless of their original format—to contribute to a single pretraining framework. The resulting representation is thus inherently universal, capable of generalizing across different skeleton domains with no structural compromises.

\begin{figure*}[t]
    \centering
    \includegraphics[width=0.7\linewidth]{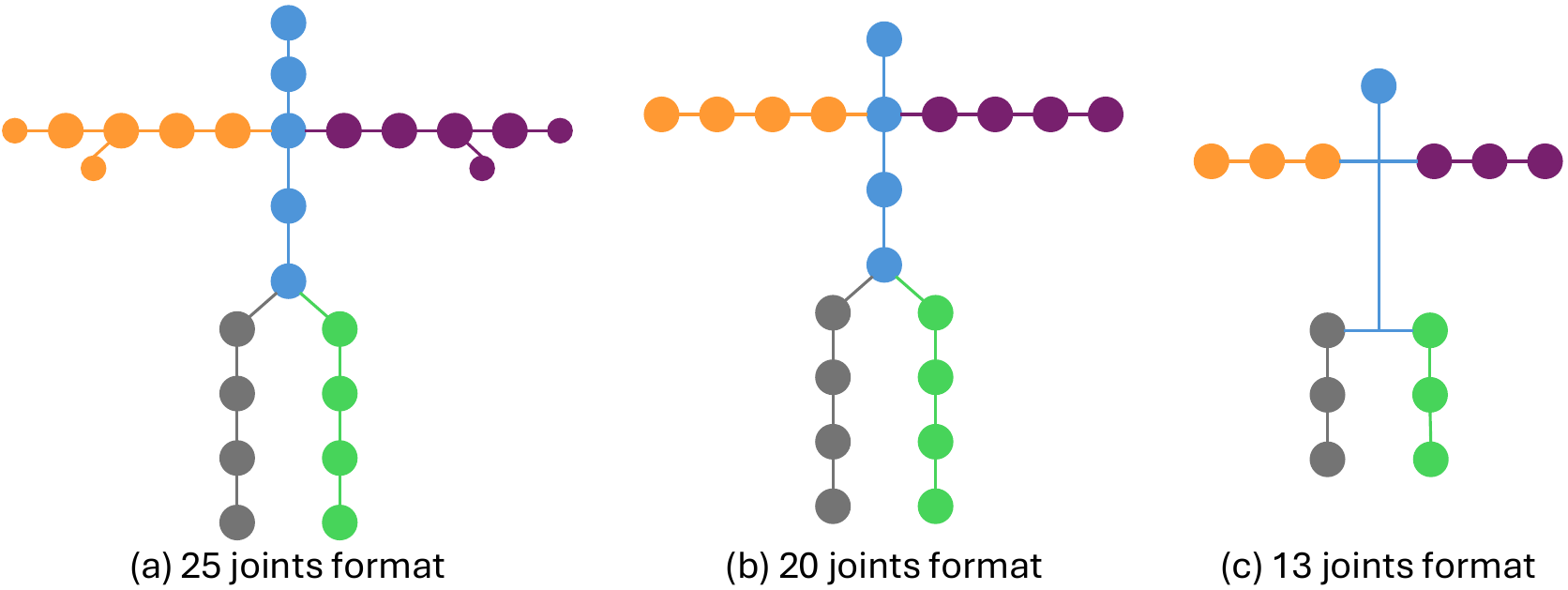}
    \caption{
    Visualization of three commonly used skeleton formats: (a) 25-joint (NTU/PKU), (b) 20-joint (NW-UCLA), and (c) 13-joint (Toyota). 
    Each skeleton is partitioned into five semantic body parts—Spine, Left Arm, Right Arm, Left Leg, and Right Leg—highlighted in different colors to ensure consistent representation across formats.
    }
    \label{fig:skeleton_compare}
\end{figure*}

\section{Universal Representation Pretraining Setups}
\label{sec: universal}

\noindent
\textbf{Implementation Details.}
To conduct the Universal Representation Pretraining experiments, we utilize the training splits from multiple datasets, including NTU120 (C-sub)~\cite{ntu120}, PKU-MMD I (CS)~\cite{pkummd}, PKU-MMD II (CS), Toyota-Smarthome (CS)~\cite{Das_2019_ICCV}, and NW-UCLA~\cite{wang2014cross}. Since the NTU60-C-sub~\cite{Shahroudy_2016_CVPR} training set is a subset of NTU120-C-sub, it is not explicitly included.

To ensure consistent input distribution across datasets, we compute the mean and standard deviation of the S2I RGB representations from the NTU120 C-sub training split, and use them to normalize the S2I inputs from all other datasets for scale alignment and stable optimization.
Additionally, due to the increased size of the combined training data, we increase the batch size from 512 to 768, while keeping all other training configurations unchanged.
All datasets are jointly trained in a unified manner, rather than using any sequential or curriculum-based strategy.

\noindent
\textbf{Skeleton Partitioning for Different Formats.}
To support universal skeleton representation learning across datasets with different joint formats, we consider three commonly used skeleton configurations, as shown in Figure~\ref{fig:skeleton_compare}. The NTU-60, NTU-120, and PKU-MMD datasets adopt the 25-joint layout extracted by Kinect V2. NW-UCLA provides 20-joint skeletons from Kinect V1, while the Toyota dataset offers 13-joint skeletons estimated using LCRNet~\cite{rogez2019lcr}.

To enable format-invariant learning, we partition the skeleton into five consistent body parts: \textit{Spine}, \textit{Left Arm}, \textit{Right Arm}, \textit{Left Leg}, and \textit{Right Leg}. Figure~\ref{fig:skeleton_compare} color-codes these body parts across the three skeleton formats, and Table~\ref{tab:partition_names} lists the corresponding joint names for each part.

\begin{table*}[t]
\centering
\caption{Semantic body part partitioning across different skeleton formats, using joint names.}
\label{tab:partition_names}
\resizebox{0.95\linewidth}{!}{
\begin{tabular}{l|p{4.2cm}|p{4.5cm}|p{4.2cm}}
\toprule
\textbf{Body Part} & \textbf{NTU/PKU (25 joints)} & \textbf{NW-UCLA (20 joints)} & \textbf{Toyota (13 joints)} \\
\midrule
Spine & head, neck, spine, middle of spine, base of spine  & head, spine, middle of spine, base of spine & head \\
\midrule
Left Arm & left shoulder, left elbow, left wrist, left hand, left thumb, tip of left hand & left shoulder, left elbow, left wrist, left hand & left shoulder, left elbow, left wrist \\
\midrule
Right Arm & right shoulder, right elbow, right wrist, right hand, right thumb, tip of right hand & right shoulder, right elbow, right wrist, right hand & right shoulder, right elbow, right wrist \\
\midrule
Left Leg & left hip, left knee, left ankle, left foot & left hip, left knee, left ankle, left foot & left hip, left knee, left ankle \\
\midrule
Right Leg & right hip, right knee, right ankle, right foot & right hip, right knee, right ankle, right foot & right hip, right knee, right ankle \\
\bottomrule
\end{tabular}
}
\end{table*}


\bibliography{sn-bibliography}